\documentclass[sigconf,nonacm]{acmart}
\renewcommand\footnotetextcopyrightpermission[1]{} 
\usepackage{natbib}
\usepackage{graphicx}
\graphicspath{ {./images/} }
\usepackage[font={footnotesize},skip=0.2pt]{caption}
\usepackage{setspace}
\usepackage{subcaption}
\usepackage{siunitx}
\newcolumntype{d}{S[
    input-open-uncertainty=,
    input-close-uncertainty=,
    parse-numbers = false,
    table-align-text-pre=false,
    table-align-text-post=false
 ]}
\AtBeginDocument{%
  \providecommand\BibTeX{{%
    \normalfont B\kern-0.5em{\scshape i\kern-0.25em b}\kern-0.8em\TeX}}}

\newcommand{\blfootnote}[1]{\begingroup\renewcommand{\thefootnote}{}\footnote{#1}\addtocounter{footnote}{-1}\endgroup}
\begin{document}

\title{Understanding Guest Preferences and Optimizing Two-sided Marketplaces: Airbnb as an Example}

\author{Yufei Wu}
\affiliation{%
  \institution{Airbnb, Inc.}
  \city{San Francisco}
  \state{California}
  \country{USA}
}
\email{yufei.wu@airbnb.com}

\author{Daniel Schmierer}
\affiliation{%
  \institution{Airbnb, Inc.}
  \city{San Francisco}
  \state{California}
  \country{USA}
}
\email{daniel.schmierer@airbnb.com}


\begin{abstract}
Airbnb is a community based on connection and belonging — many hosts on Airbnb are everyday people who share their worlds to provide guests with the feeling of connection and being at home; Airbnb strives to connect people and places. Among our efforts to connect guests and hosts, we provide tools to enable hosts to set competitive prices, which helps improve affordability for guests while helping hosts get more bookings. We also personalize the guest experience to show them the listings that match their needs. 

To help inform these efforts, we combine economic modeling and causal inference techniques to understand how guests book stays based on the prices hosts set, among other factors, and how that preference varies across different guests and listings. Such understanding helps us identify opportunities for Airbnb to support the marketplace and better connect guests and hosts. For example, understanding how much guests respond to different prices helps optimize the tools that we provide to hosts, in order to enable hosts to choose and set competitive prices that further balance demand and supply. As another example, understanding heterogeneity in guest preferences helps us personalize the guest experience and better match them with the listings that meet their needs, based on how much they respond to different prices and other factors. 
\end{abstract}


\maketitle
\blfootnote{\textcopyright\ 2024 Airbnb, Inc. All rights reserved.}

\section{Introduction} \label{Intro}
Airbnb is a community based on connection and belonging—a community that was born in 2007 when two Hosts welcomed three guests to their San Francisco home, and has since grown to over 5 million Hosts who have welcomed over 1.5 billion guest arrivals to over 100,000 cities and towns in almost every country and region across the globe. Hosts on Airbnb are everyday people who share their worlds to provide guests with the feeling of connection and being at home. We strive to connect people and places.

For two-sided marketplaces, pricing is generally an important lever to balance supply and demand. But Airbnb does not set the prices that guests pay. Instead, hosts set prices for their places, while Airbnb sets service fees, which are a percentage of the value of a booking. There are several reasons why it can be challenging for hosts to know what is the right price to set for their place. First, the listings on Airbnb are highly heterogeneous, in terms of location, home design, and amenities, making the pricing problem unique for each host.  Second, some hosts may have limited experience hosting. Finally, because travel and accommodation is not an everyday occurrence, bookings may come only relatively infrequently, providing hosts with little feedback about the appropriateness of prices.

As part of our efforts to connect guests and hosts, we provide tools that help hosts to choose and set competitive prices, in order to best meet the demand from guests. By providing the tools, Airbnb can not only provide better value to guests but also help hosts attract more business. For example, \citet*{ye2018customized}, a prior paper that Airbnb’s Pricing Modeling team presented at KDD, discusses a predictive pricing model that empowered price suggestions to help hosts choose and set competitive prices. We also personalize guest experience and host tools to better match guests with the listings that tailor to their needs. For instance, \citet*{10.1145/3109859.3109920} and \citet*{10.1145/3219819.3219885}, two papers that Airbnb’s Relevance team presented at KDD, explain how Airbnb personalizes search ranking to match guests with the listings that meet their short-term and long-term interests. As another example, an Airbnb Blog Post by \citet*{jing2023} discusses how Airbnb leverages machine learning models to personalize recommendations to hosts based on guest preferences for different home attributes, such as acquiring the amenities that guests are looking for the most. 

To help inform these efforts, we combine economic modeling with causal inference techniques to understand how guests book stays based on the prices hosts set, among other factors, and how that preference varies across different guests and listings. Such understanding helps us identify opportunities for Airbnb to optimize the marketplace and better connect guests and hosts. Specifically, we measure how sensitive guests are to prices, to help optimize host pricing tools, enabling hosts to choose and set competitive prices and better meet guest demand. 

We also try to understand heterogeneity in guest preferences, or how the preference for affordability varies across different segments of guests. Such insights identify opportunities for guest experience personalization, to help them more easily find the listings that meet their needs, based on their preference for affordability and other factors. 

\section{Experimental vs Observational Methods} \label{MethodComparison}
Both experimental and non-experimental methods (based on observational data) are useful and widely applied to understand consumer preferences, including how much they value affordability, or how sensitive they are to price changes. 

Experiments have a long history in online marketplaces and, when done correctly, are still seen as the gold standard for causal inference. However, there are several nuances when running experiments around pricing, which have been discussed by data scientists and economists at Airbnb as well as other tech companies (see, for example, \citet*{holtz2020reducing}, \citet*{johari2022experimental}, and \citet*{le2023price}). An important concern for marketplace experiments is interference, that an intervention on a market participant would affect others. While pricing experiments can be designed to minimize bias from interference, those designs come at the cost of statistical power. Further constraints on statistical power include the limited ability to introduce price variation where the marketplace does not control prices directly, and infrequent participation in the marketplace. 

Given these nuances around pricing experimentation, observational methods are also very useful for understanding guest price sensitivity in a marketplace like Airbnb. There is more variation in prices in observational data, across listings and over time, compared to pricing experiments. We draw on the academic literature in economics, which has historically been focused on drawing inference from observational data. We combine that with experimental studies to validate and calibrate the estimates from the observational method, to address the general challenge with establishing unbiasedness of results for observational analysis. 

\section{Understand Guest Price Sensitivity to Enable Hosts to Set Competitive Prices} \label{Methodology}
How do we model guest demand and measure guest price sensitivity using observational data? Our method is grounded in the economic theory of guest and host behavior. Specifically, we begin by modeling guest behavior to inform our empirical problem. We then leverage marketplace dynamics, considering the interaction between host supply and guest demand, to isolate the impact of price, or affordability, on guest demand. 

\subsection{Model guest choice}
Following a well-established strand of academic economics literature, we model guest choice between different listing options using a logit model. The model assumes that potential guest $i$’s utility, or overall satisfaction, from choosing a certain product $j$ in time $t$ can be expressed as follows, where $p$ denotes price, $X$ denotes observed product attributes, and $\xi$ represents utility from unobserved attributes of the product.\footnote{For model tractability, we group listings based on their geography and selected attributes into products, and calculate product price and attributes as the average across listings. A common assumption in such models is that the utility from a product only depends on its own characteristics and not those of other products. We adopt this assumption here as well.} The error term $\epsilon$ follows Type I Extreme Value distribution.\footnote{This error distribution is commonly assumed in demand modeling to generate tractable substitution patterns.}
\begin{equation}\label{eq:utility}
 u_{ijt} = - \alpha p_{jt} + X^T_{jt}\beta + \xi_{jt} + \epsilon_{ijt} 
\end{equation}
A guest will choose the product that provides the highest utility out of all available options. As a result, the share of guests choosing product $j$, as the aggregated choice probability across guests, is as follows. Intuitively, the higher the price hosts set, the lower share of guests will choose the product, holding everything else the same. 
\begin{equation}\label{eq:share}
s_{jt} = \frac{exp(\delta_{jt})}{1+\sum^J_{j'=1} exp(\delta_{j't})}
\end{equation}
where $\delta_{jt}$ represent the mean utility from product $j$ in $t$:
\begin{equation}\label{eq:delta}
\delta_{jt} = - \alpha p_{jt} + \boldsymbol X^T_{jt}\boldsymbol\beta + \xi_{jt} 
\end{equation}
The price parameter $\alpha$ quantifies how sensitive guests are to host price changes. To interpret it in a standardized unit-free manner, we use the concept of the price elasticity of demand, which tells us if hosts were to drop (or raise) price by 1\%, how much bookings would increase (or decrease). This price elasticity of demand can be derived as a function of the estimated price parameter $\hat{\alpha}$, the average price $\bar{p}$, and the average product share $\bar{s}$.
\begin{equation}\label{eq:elasticity}
\hat e = -\hat \alpha \bar p  \left( 1-\bar s\right)
\end{equation}

We follow the method developed by \citet*{berry1995auto} (aka “BLP”) to simplify the estimation problem, to simplify the empirical estimation into a linear regression problem and apply instrumental variables to isolate the causal impact of price. The next section will discuss how we construct instrumental variables to help isolate causality from observational data. 

\subsection{Empirically estimate guest price sensitivity}
Observational data provides rich information on how guests tend to choose their stays, when they have different options at a range of different prices. These signals help us estimate the guest choice model above and understand how guests respond to changes in host prices. However, unlike in experiments, real-world price variation is endogenous rather than random, creating a challenge for estimating the guest price elasticity.  

\begin{figure}[h]
  \caption{Supply and demand jointly determine equilibrium price}
  \centering
  \includegraphics[width=0.8\linewidth]{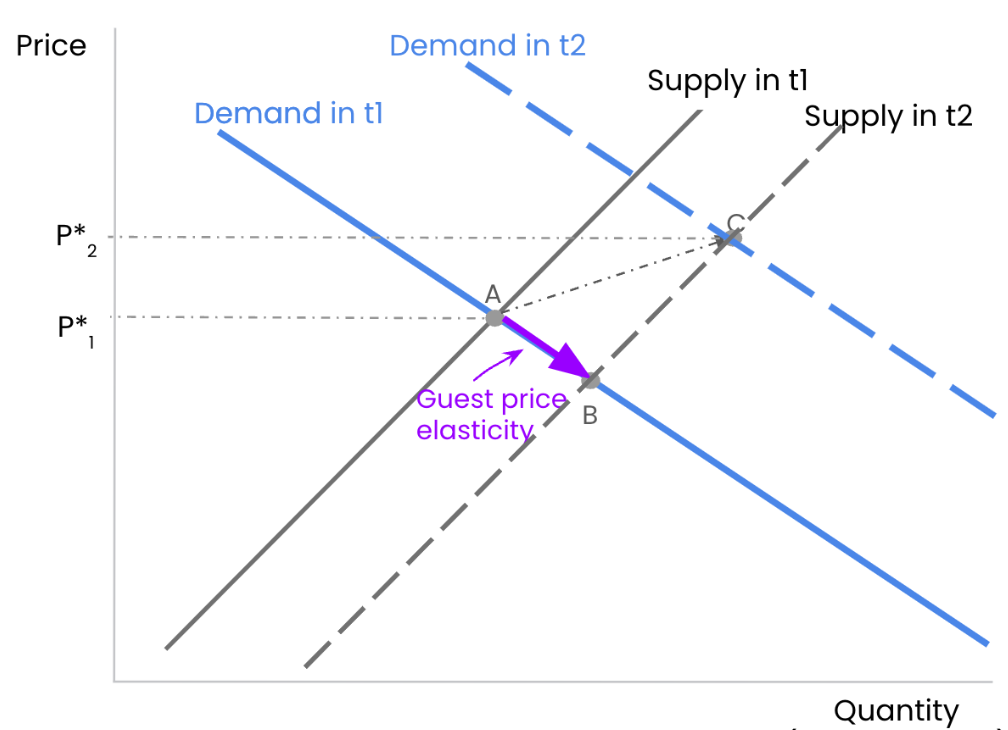}
  \label{fig:supplydemand}
\end{figure}

Figure \ref{fig:supplydemand} visualizes the demand curve (describing how many nights guests would book at each price level) and the supply curve (describing how many listing nights hosts would supply at each price level) for Airbnb stays across two observation periods, t1 and t2. We aim to measure guest price elasticity, or how guest demand responds to price changes while holding everything else unchanged — that is the slope of the demand curve, measurable by comparing outcomes A and B. In reality, guest demand and host supply jointly determine the equilibrium price and quantity we observe. Thus, we would be comparing outcomes A and C instead if we take observational data from t1 and t2 at face value. This comparison mixes two types of quantity changes — guests’ demand response to a supply-driven price change (the change from A to B; the lower the price, the more nights guests would book), and hosts’ supply response to a demand-driven price change (the change from B to C; the higher the price, the more listing nights hosts would supply).

\begin{figure}[h]
  \centering
  \caption{Heat Maps of Channel Residual Impressions Across DMAs Over Time}
    \begin{subfigure}[a]{0.5\textwidth}
        \centering
        \caption{Motivating Example of Two Geos}
        \includegraphics[width=\linewidth]{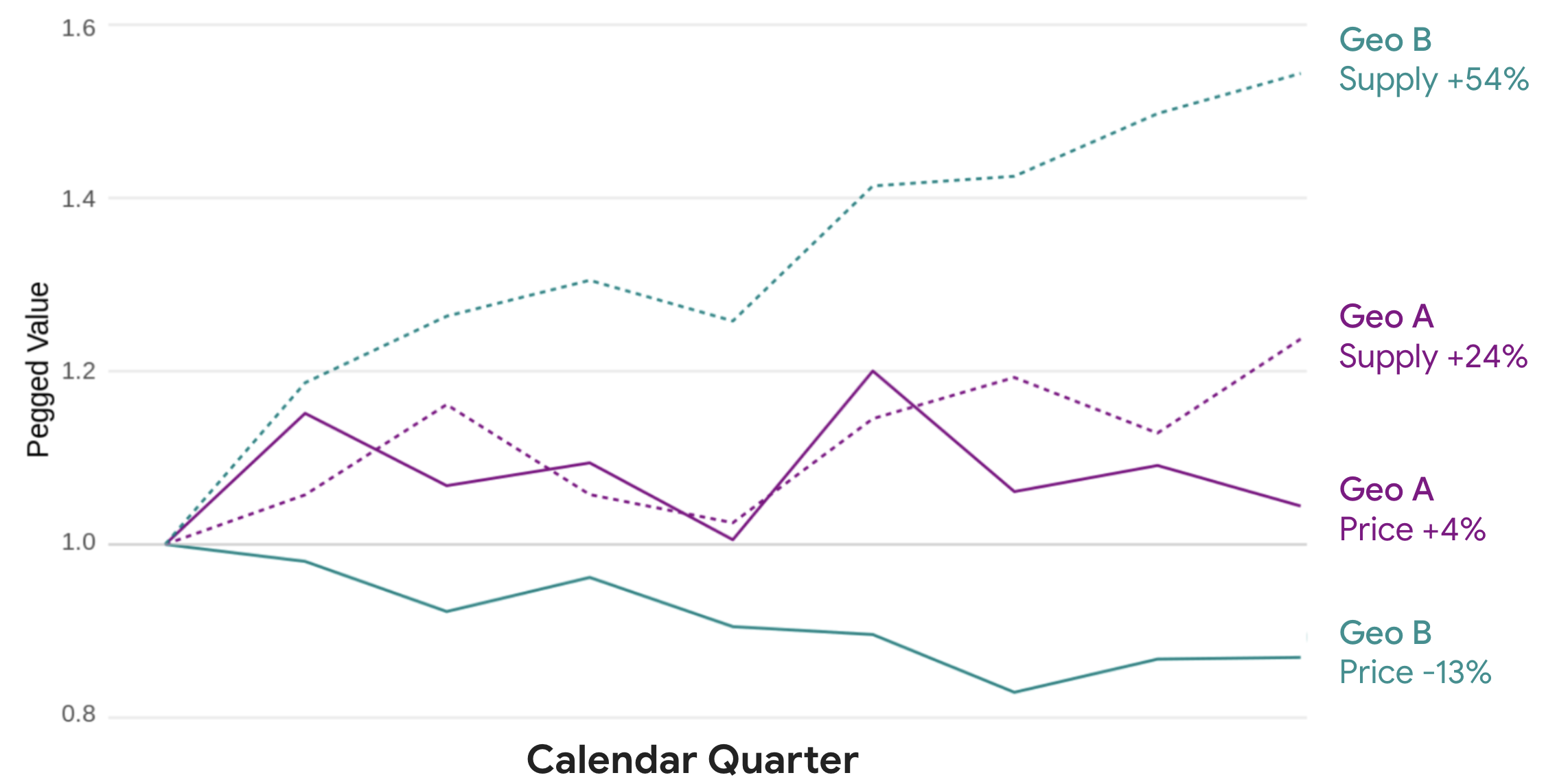}   
    \end{subfigure}
    \hfill
    \begin{subfigure}[b]{0.5\textwidth}
        \centering
        \caption{Generalizing to All Geographies}
        \includegraphics[width=\linewidth]{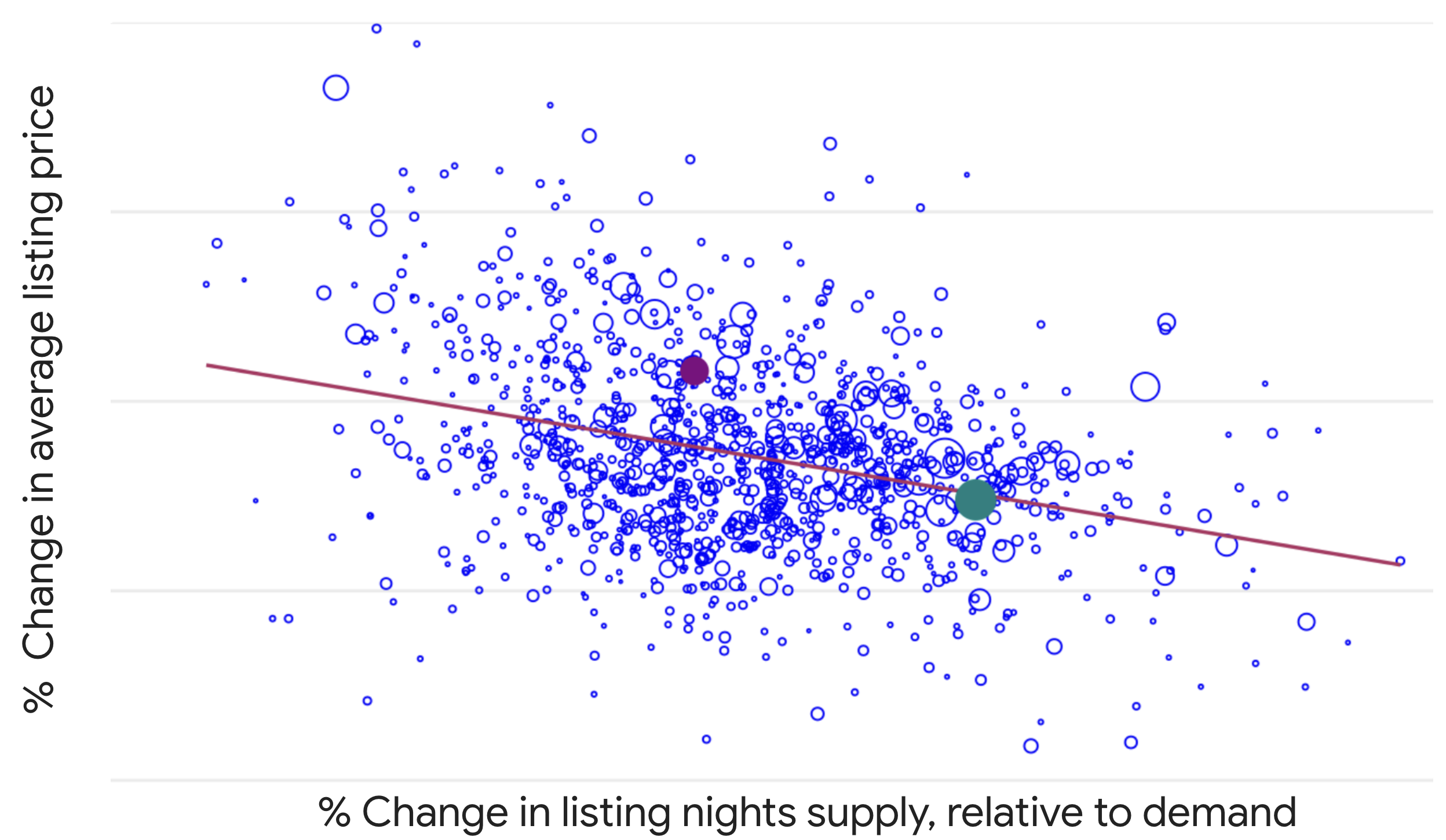} 
    \end{subfigure}
 \label{fig:variation_new}
\end{figure}

To isolate the guest price elasticity of interest (the slope between A and B in Figure 1), we utilize methods that draw on sources of variation that are plausibly exogenous to demand. We focus on price variation induced by differences in supply relative to demand across geographies. As a motivating example, Figure \ref{fig:variation_new}(A) compares two popular destination geos for Airbnb stays, demonstrating how differential supply growth could potentially introduce an exogenous source of price variation. The horizontal axis represents calendar quarter, while the vertical axis represents supply and price, both normalized to one at the beginning of the example period. Over this time period, average price increased 4\% in Geo A but decreased 13\% in Geo B. Part of that difference was driven by the fact that Geo B has experienced faster supply growth than Geo A. 

This example compares supply growth without considering possible demand growth, begging the question of whether such supply variation could be a response to demand. Next, we control for the level of demand and focus on supply variation that’s driven by differences in factors other than demand, such as heterogeneity in short term rental regulations and cost of providing accommodation across geographies. As \ref{fig:variation_new}(B) illustrates, the comparison between Geo A and Geo B is robust to accounting for the realized level of demand, as measured by the total number of listing views. The horizontal axis represents listing supply growth over a certain time period, conditional on realized demand, while the vertical axis represents changes in the average price over the same time period. Relative to realized demand, listing supply has grown by over 60\% in Geo B (the green dot) but has grown much less in Geo A (the purple dot) over the same time period, explaining why prices decreased more in Geo B relative to Geo A. This relationship generalizes to other destination geos, as shown by the blue circles.\footnote{Each circle represents a destination geo, with the circle size scaled according to its level of realized demand.} There is a clear negative relationship between price and supply, conditional on demand, as illustrated by the red fitted line from a linear regression. The faster listing supply grows relative to demand, the lower the prices that hosts set on average. 

This empirical relationship motivates our instrumental variables approach. Controlling for the level of realized demand, we use differential supply growth across geographies as an instrument for price variation, to isolate supply-driven price changes and estimate guest price elasticity. To the extent that hosts adjust listing supply without perfect foresight of demand changes, variation in listing supply induces plausibly exogenous price variation, which we use to estimate the guest choice model and understand the causal impact of price on guest bookings.  

\subsection{Use experimental results to validate and calibrate the model}
We acknowledge a general limitation with observational methods: it is difficult to establish unbiasedness of the results without a randomized experiment. While we use instrumental variables to help isolate the causal impact of price, the unbiasedness of the results depends on the validity of the exclusion restriction (see, for example, Chapter 5 of \citet*{wooldridge2010econometric}), which assumes that our instrument affects guest choice only through price and not by any other means. However, proving the validity of such exclusion restrictions is not possible. To address this challenge and to ensure confidence in our findings, we leverage experimental results to validate and calibrate our model, to the extent possible. 

When we initially developed this method, we relied on carefully designed pricing experiments that minimized interference bias as the source of ground truth. We estimated our model using data from around the same time as these experiments and compared our estimates to experimental results. These experimental results serve two purposes. First, they help qualitatively confirm the findings from our model, particularly regarding guest preference heterogeneity. For instance, segments estimated to be particularly sensitive to prices in our model are also found to be sensitive to prices in the experimental data. Second, while our model estimates are in the same range as the experiment estimates, comparing them helps us assess the direction and degree of bias from our observational method. Specifically, our estimates of guest price sensitivity are slightly higher in magnitude than experimental results. We correct for this upward bias by applying a haircut based on the difference between our model estimates and the experimental results, using data from a similar time period. 

Through this validation and calibration process, we are able to update our model estimates using more recent data. We also incorporate new experimental data as it becomes available. In fact, after initially calibrating the model to historial experiments and applying to more recent data, a subsequent experiment further validated the findings, showing that our latest estimates closely matched experimental results. This demonstrates the value of continuing to use experiment data to validate observational methods, reinforcing our confidence in the insights derived from those methods. 

In summary, our observational method and experimentation complement rather than substitute each other. While our method provides ongoing measurement and unlocks insights at more granular levels by leveraging more data, experimentation helps validate our findings and calibrate our estimates to adjust for any potential bias. 

\subsection{Provide on-going measurement to improve affordability and host success }
In summary, in addition to using experiments, it is valuable to apply choice models from the economics literature to non-experimental data, and leverage supply-driven price variation to measure guest price sensitivity. Validated against experimental data, our model enhances our ability to measure guest price sensitivity and understand guest preferences on an ongoing basis using observational data. 

This capability allows us to detect important trends in guest preferences and leverage those insights to continually optimize the tools we provide to hosts. As discussed earlier, it can be difficult for hosts to determine the right price to set for their place because listings are highly heterogeneous, their hosting experience varies, and the relatively infrequent nature of bookings limits timely feedback for host pricing decisions. Additionally, guest preferences can change over time, making it even more challenging for host-set prices to stay competitive without any guidance. By understanding guest price sensitivity on an ongoing basis and monitoring important changes over time, we can update our tools to better assist hosts with their pricing strategies, both to accommodate the evolving needs of guests and to help hosts optimize for success. 

\section{Understand Preference Heterogeneity to Personalize Guest Experience}
As discussed earlier, we follow a well established strand of economics literature and model guest choice using a logit model. The logit model is widely used because it provides tractable product shares and substitution patterns, simplifying the empirical estimation. However, a technical nuance of a logit model is that it predicts a strictly positive share for every product. That means the model’s predicted shares cannot match the observed product shares for some observations, creating a challenge for estimation, especially when we consider smaller guest segments. As \citet*{gandhi2021empirical} points out, dropping product observations with zero shares or assigning them small positive shares can result in large biases. \citet*{dube2021random} and \citet*{gandhi2023estimating} developed two methods to address this challenge, but these methods either require additional assumptions or provide only partial parameter identification. 

We propose a solution to this challenge that is relatively easy to implement and interpret, which addresses our need for understanding preference heterogeneity by guest segment. Specifically, we first model the theoretical relationship between (A) how the guest mix changes in response to price changes; and (B) the heterogeneity of guest price elasticity across guest segments.  As formularized in Equation (\ref{eq:relationship}), how a guest segment k’s share of conversions ($\sigma^k$)  responds to changes in price (p) is determined by segment k’s price elasticity ($e^k$), relative to that of a representative guest ($\bar{e}$). 

\begin{equation}\label{eq:relationship}
\frac{\partial \ln(\sigma^{k}+1)}{\partial \ln(p)} = (e^{k} - \bar{e})\frac{\sigma^k}{\sigma^k+1} 
\end{equation}

We then empirically understand this theoretical relationship, by estimating how the guest mix responds to price changes using the following panel linear regression. We apply the supply-based instrumental variables for price ($p$) to causally estimate $\eta$, the sensitivity of guest mix to price changes. 
\begin{equation}\label{eq:plm}
\ln (\sigma^{k}_{jt} + 1) = \theta_t + \eta \ln(p_{jt}) +  \boldsymbol X^T_{jt} \boldsymbol\gamma + \epsilon_{jt}
\end{equation}

\begin{figure}[h]
  \centering
  \caption{Hypothetical illustration of preference heterogeneity}
  \includegraphics[width=\linewidth]{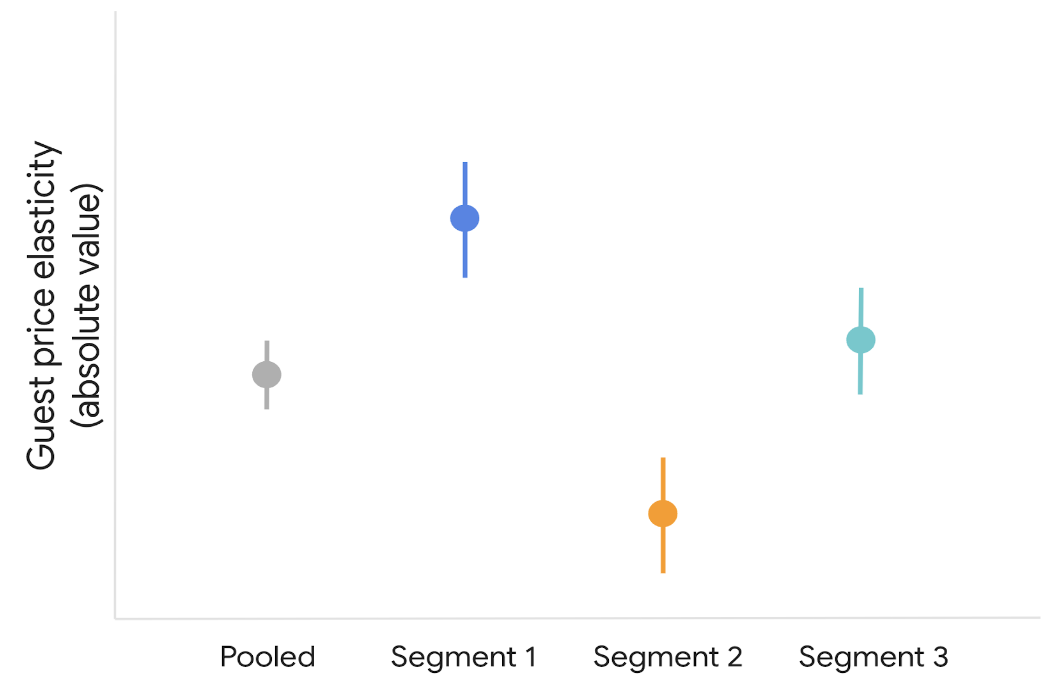}
  \label{fig:example}
\end{figure}

After estimating $\hat{\eta}$ from equation (\ref{eq:plm}), we can combine that with the overall price elasticity estimate  e,  to derive the price elasticity for different guest segments. Such insights, combined with other findings on heterogeneous preferences for listing attributes, provides guidance on improving the personalization of experience for different segments of guests. Using the hypothetical example in Figure \ref{fig:example} for illustration, suppose guests in segment 1 values affordability more than other guests (i.e., they have higher price elasticity), but values certain listing attributes less (e.g., high ratings). That suggests opportunities for helping guests in segment 1 find more affordable listings more easily. One option could be to curate marketing communications for these guests to highlight affordable listings in the destinations they have recently searched, which are more likely to meet their needs, compared to more expensive listings that may be appealing in other aspects, such as high ratings. While this serves as a simplified hypothetical example, the actual product iterations are more complex and require careful consideration.

\section{Conclusion} \label{conclusion}
In this paper, we combine economic modeling with causal inference methods, in order to understand guest preferences and measure guest price elasticity using observational data. This approach complements experiments, which establish the ground truth for calibration but are subject to low statistical power and practical constraints. Our method not only enables ongoing measurement but also helps identify preference heterogeneity across guests. As a result, we can pinpoint opportunities to optimize host tools and personalize guest experience, enhancing our ability to connect guests and hosts.

While well-designed experimentation remains the gold standard for measurement, our research contributes to enhancing the feasibility and continuity of understanding consumer preferences in two-sided marketplaces, and leveraging those insights for ongoing product optimization. Our approach is particularly valuable in settings where it is particularly difficult to experiment, such as when the marketplace has limited direct control over prices. For instance, our method can be readily applied to study consumer preferences on consumer-to-consumer (“C2C”) e-commerce marketplaces, where sellers list products and set prices, buyers make purchasing decisions, and the e-commerce marketplace charges fees as a percentage of seller-set prices. For such platforms, understanding consumer preferences and heterogeneity can generate important insights to help sellers set competitive prices and improve products to meet consumer needs and build successful businesses; it can also provide insights to help match consumers with the product offerings that meet their needs. Our method allows for model consumer choices among alternative products and using instrumental variables to isolate the causal impact of price and product features on buyer choice. In our setting, we construct supply-based instruments to isolate the causal impact of price, but with proper instruments, the approach can also help isolate the causal impact of other product features.

In conclusion, our approach bridges the gap between experimental and observational methods, providing a robust framework for ongoing understanding of guest preferences in two-sided marketplaces. It can be easily adapted to other two-sided marketplaces facing similar challenges to pricing experimentation, such as C2C e-commerce platforms. 

\section{Acknowledgement}
The authors would like to thank Amit Gandhi (Wharton and Airbnb, Inc.), Alex Deng (Microsoft), Nick Riabov (Airbnb, Inc.), Navin Sivanandam (Airbnb, Inc.), Mitra Akhtari (Airbnb, Inc.), Brian Weller (Airbnb, Inc.), Shanni Weilert (Airbnb, Inc.), and Ali Rauh (Uber Technologies, Inc.) for their contributions and support for this project.

\nocite{berry1995auto}
 

\bibliography{sample-base}

\begin{thebibliography}{12}
\providecommand{\natexlab}[1]{#1}
\providecommand{\url}[1]{\texttt{#1}}
\expandafter\ifx\csname urlstyle\endcsname\relax
  \providecommand{\doi}[1]{doi: #1}\else
  \providecommand{\doi}{doi: \begingroup \urlstyle{rm}\Url}\fi

\bibitem[Ye et~al.(2018)Ye, Qian, Chen, Wu, Zhou, De~Mars, Yang, and Zhang]{ye2018customized}
Peng Ye, Julian Qian, Jieying Chen, Chen-hung Wu, Yitong Zhou, Spencer De~Mars, Frank Yang, and Li~Zhang.
\newblock Customized regression model for airbnb dynamic pricing.
\newblock In \emph{Proceedings of the 24th ACM SIGKDD international conference on knowledge discovery \& data mining}, pages 932--940, 2018.

\bibitem[Grbovic(2017)]{10.1145/3109859.3109920}
Mihajlo Grbovic.
\newblock Search ranking and personalization at airbnb.
\newblock In \emph{Proceedings of the Eleventh ACM Conference on Recommender Systems}, RecSys '17, page 339–340, New York, NY, USA, 2017. Association for Computing Machinery.
\newblock ISBN 9781450346528.
\newblock \doi{10.1145/3109859.3109920}.
\newblock URL \url{https://doi.org/10.1145/3109859.3109920}.

\bibitem[Grbovic and Cheng(2018)]{10.1145/3219819.3219885}
Mihajlo Grbovic and Haibin Cheng.
\newblock Real-time personalization using embeddings for search ranking at airbnb.
\newblock In \emph{Proceedings of the 24th ACM SIGKDD International Conference on Knowledge Discovery \& Data Mining}, KDD '18, page 311–320, New York, NY, USA, 2018. Association for Computing Machinery.
\newblock ISBN 9781450355520.
\newblock \doi{10.1145/3219819.3219885}.
\newblock URL \url{https://doi.org/10.1145/3219819.3219885}.

\bibitem[Jing and Xia(2023)]{jing2023}
Joy Jing and Jing Xia.
\newblock Prioritizing home attributes based on guest interest.
\newblock \emph{the Airbnb Tech Blog}, 2023.

\bibitem[Holtz et~al.(2020)Holtz, Lobel, Liskovich, and Aral]{holtz2020reducing}
David Holtz, Ruben Lobel, Inessa Liskovich, and Sinan Aral.
\newblock Reducing interference bias in online marketplace pricing experiments.
\newblock \emph{arXiv preprint arXiv:2004.12489}, 2020.

\bibitem[Johari et~al.(2022)Johari, Li, Liskovich, and Weintraub]{johari2022experimental}
Ramesh Johari, Hannah Li, Inessa Liskovich, and Gabriel~Y Weintraub.
\newblock Experimental design in two-sided platforms: An analysis of bias.
\newblock \emph{Management Science}, 68\penalty0 (10):\penalty0 7069--7089, 2022.

\bibitem[Le and Deng(2023)]{le2023price}
Thu Le and Alex Deng.
\newblock The price is right: Removing a/b test bias in a marketplace of expirable goods.
\newblock In \emph{Proceedings of the 32nd ACM International Conference on Information and Knowledge Management}, pages 4681--4687, 2023.

\bibitem[Berry et~al.(1995)Berry, Levinsohn, and Pakes]{berry1995auto}
Stephen Berry, James Levinsohn, and Ariel Pakes.
\newblock Automobile prices in market equilibrium.
\newblock \emph{Econometrica}, 63\penalty0 (4):\penalty0 841--890, 1995.

\bibitem[Wooldridge(2010)]{wooldridge2010econometric}
Jeffrey~M Wooldridge.
\newblock \emph{Econometric analysis of cross section and panel data}.
\newblock MIT press, 2010.

\bibitem[Gandhi and Nevo(2021)]{gandhi2021empirical}
Amit Gandhi and Aviv Nevo.
\newblock Empirical models of demand and supply in differentiated products industries.
\newblock In \emph{Handbook of industrial organization}, volume~4, pages 63--139. Elsevier, 2021.

\bibitem[Dub{\'e} et~al.(2021)Dub{\'e}, Horta{\c{c}}su, and Joo]{dube2021random}
Jean-Pierre Dub{\'e}, Ali Horta{\c{c}}su, and Joonhwi Joo.
\newblock Random-coefficients logit demand estimation with zero-valued market shares.
\newblock \emph{Marketing Science}, 40\penalty0 (4):\penalty0 637--660, 2021.

\bibitem[Gandhi et~al.(2023)Gandhi, Lu, and Shi]{gandhi2023estimating}
Amit Gandhi, Zhentong Lu, and Xiaoxia Shi.
\newblock Estimating demand for differentiated products with zeroes in market share data.
\newblock \emph{Quantitative Economics}, 14\penalty0 (2):\penalty0 381--418, 2023.

\end{thebibliography}


\end{document}